%% file: main.tex
\documentclass[sigconf, screen, authorversion, nonacm, acmnumeric, natbib=false]{acmart}
\AtBeginDocument{%
  \providecommand\BibTeX{{%
    \normalfont B\kern-0.5em{\scshape i\kern-0.25em b}\kern-0.8em\TeX}}}

\input{math-symbols}

\newcommand{\secref}[1]{sec.~\ref{sec:#1}}
\newcommand{\figref}[1]{fig.~\ref{fig:#1}}
\newcommand{\tabref}[1]{tab.~\ref{tab:#1}}

\newcommand{\Figref}[1]{Fig.~\ref{fig:#1}}

\usepackage{todonotes}
\newcommand{\minisec}[1]{\par\smallskip\noindent\textbf{#1.}}
\usepackage{siunitx}
\usepackage{booktabs}
\usepackage{multirow}
\usepackage{pifont} %
\usepackage{subcaption}

\DeclareSIUnit{\million}{\text{M}}
\RequirePackage[
datamodel=acmdatamodel,
style=acmnumeric, %
]{biblatex}
\begin{document}
\title{Weight Sparsity Complements Activity Sparsity in Neuromorphic Language Models}
\author{Rishav Mukherji}
\authornote{Both authors contributed equally to the paper}
\authornote{Work carried out at TUD Dresden University of Technology}
\affiliation{%
        \institution{Birla Institute of Technology and Science Pilani -- Goa Campus}
        \city{Goa}
        \country{India}
}
\author{Mark Sch{\"o}ne}
\authornotemark[1]
\affiliation{%
    \institution{TUD Dresden University of Technology}
    \city{Dresden}
    \country{Germany}
}
\email{mark.schoene@tu-dresden.de}
\author{Khaleelulla Khan Nazeer}
\affiliation{%
    \institution{TUD Dresden University of Technology}
    \city{Dresden}
    \country{Germany}
}
\author{Christian Mayr} 
\affiliation{%
    \institution{Center for Scalable Data Analytics and Artificial Intelligence (ScaDS.AI)}
    \city{}
    \country{}
}
\affiliation{%
    \institution{Centre for Tactile Internet with Human-in-the-Loop (CeTI)}
    \city{}
    \country{}
}
\affiliation{%
    \institution{TUD Dresden University of Technology}
    \city{Dresden}
    \country{Germany}
}
\author{David Kappel}
\affiliation{%
    \institution{Ruhr University Bochum}
    \city{Bochum}
    \country{Germany}
}
\author{Anand Subramoney}
\affiliation{%
    \institution{Royal Holloway, University of London}
    \city{London}
    \country{United Kingdom}
}
\renewcommand{\shortauthors}{Mukherji and Sch{\"o}ne et al.}
\begin{abstract}
Activity and parameter sparsity are two standard methods of making neural networks computationally more efficient.
Event-based architectures such as spiking neural networks (SNNs) naturally exhibit activity sparsity, and many methods exist to sparsify their connectivity by pruning weights.
While the effect of weight pruning on feed-forward SNNs has been previously studied for computer vision tasks, the effects of pruning for complex sequence tasks like language modeling are less well studied since SNNs have traditionally struggled to achieve meaningful performance on these tasks.
Using a recently published SNN-like architecture that works well on small-scale language modeling, we study the effects of weight pruning when combined with activity sparsity.
Specifically, we study the trade-off between the multiplicative efficiency gains the combination affords and its effect on task performance for language modeling.
To dissect the effects of the two sparsities, we conduct a comparative analysis between densely activated models and sparsely activated event-based models across varying degrees of connectivity sparsity.
We demonstrate that sparse activity and sparse connectivity complement each other without a proportional drop in task performance for an event-based neural network trained on the Penn Treebank and WikiText-2 language modeling datasets.
Our results suggest sparsely connected event-based neural networks are promising candidates for effective and efficient sequence modeling.
\end{abstract}
\maketitle
\input{content/introduction}
\input{content/related-work}
\input{content/methods}
\input{content/results}
\input{content/discussion}
\begin{acks}
    Rishav Mukherji was funded by the DAAD WISE scholarship for the duration of the work.
    Mark Schöne is supported with funds from Bosch-Forschungsstiftung im Stifterverband.
    Khaleelulla Khan Nazeer is funded by the German Federal Ministry of Education and Research (BMBF), funding reference 16ME0729K, joint project "EVENTS".
    This work was partially funded by the German Federal Ministry of Education and Research (BMBF) and the free state of Saxony within the ScaDS.AI center of excellence for AI research.
    David Kappel is funded by the German Federal Ministry for Economic Affairs and Climate Action (BMWK) project ESCADE (01MN23004A).
    Christian Mayr is affiliated to German Research Foundation (DFG, Deutsche Forschungsgemeinschaft) as part of Germany’s Excellence Strategy – EXC 2050/1 – Project ID 390696704 – Cluster of Excellence “Centre for Tactile Internet with Human-in-the-Loop” (CeTI) of Technische Universität Dresden.  
    The authors gratefully acknowledge the computing time made available to them on the high-performance computer at the NHR Center of TU Dresden. This center is jointly supported by the Federal Ministry of Education and Research and the state governments participating in the NHR (www.nhr-verein.de/unsere-partner).
\end{acks}

\printbibliography
\end{document}

%% file: math-symbols.tex
\usepackage{graphicx}

\newcommand{\bfa}{\mathbf{a}}
\newcommand{\bfu}{\mathbf{u}}
\newcommand{\bfr}{\mathbf{r}}
\newcommand{\bfs}{\mathbf{s}}

\newcommand{\bfc}{\mathbf{c}}
\newcommand{\bfx}{\mathbf{x}}
\newcommand{\bfy}{\mathbf{y}}
\newcommand{\bfz}{\mathbf{z}}

\newcommand{\bfth}{\boldsymbol{\vartheta}}

\newcommand{\bfW}{\ensuremath{\mathbf{W}}}

\newcommand{\bfb}{\ensuremath{\mathbf{b}}}

\newcommand{\dti}{^{\langle t \rangle}}
\newcommand{\dtim}[1]{^{\langle t-{#1} \rangle}}

\newcommand{\hp}{{\,\mathrel{\vcenter{\hbox{\tiny$\odot$}}}\,}}

\newcommand{\bwrap}[1]{\left( #1 \right)}

\newcommand{\fun}[2]{{#1}\bwrap{#2}}

%% file: content/introduction.tex
\section{Introduction}
Machine learning methods have become increasingly popular for applications ranging from cloud services to mobile and edge systems due to the steady increase in available compute.
While task performance is crucial for all applications, energy consumption is particularly critical for deployment in environments such as robotics or mobile devices.
Many tasks have additional latency requirements to maintain safety demands or to enhance the user experience.

To meet the computational demands and constrains of future machine learning systems, neuromorphic computing pursues a co-design process of biologically plausible neural networks and learning methods, along with efficient silicon implementations of these systems.
Biologically plausible spiking neural networks (SNNs) have found a wide range of applications in the machine learning domains of vision \cite{Liu2022, Fang2023}, audio \cite{hammouamri2024learning, Bittar2022}, and robotics \cite{zhao2020neuromorphic}.
These models address the energy issue of deep learning by implementing the sparse communication paradigm of biological neural networks.

\begin{figure*}[!ht]
    \centering
    \includegraphics[width=\textwidth]{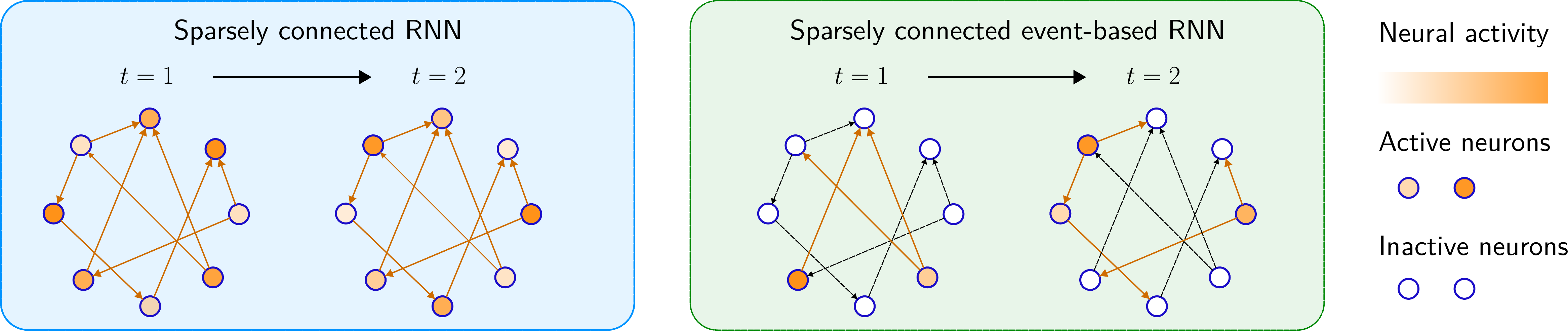}
    \caption{
        Sparsely connected artificial neural networks (ANNs) such as LSTMs \cite{Hochreiter1997} transfer their entire state for all simulation steps, while event-based neural networks such as EGRU \cite{Subramoney2023} only transfer a fraction of their state in each step.
    }
    \label{fig:figure1}
\end{figure*}

Besides sparse event-based communication, 
sparse network connectivity is a common method to reduce the computational footprint of neural networks.
Joint activity sparsity and connectivity sparsity has the potential to further reduce the overall communication within neural networks, as illustrated  in \figref{figure1}.
But the joint effect of these two sparsities on task performance is not well understood, especially for a complex sequence task such as language modeling.

This is because, while biologically plausible SNNs excel in static computer vision applications such as image classification~\cite{Liu2022, Fang2023}, they do not achieve meaningful performance in complex sequence tasks such as language modeling.
To remedy this, several authors have proposed mutually related models that are inspired by the sparse and event-based communication protocol of biological neural networks without detailed biological plausibility \cite{Neil2017, connor2017sigma,Wozniak2020, Rezaabad2020, Subramoney2023, shen2023astrocyteenabled}.
These ``event-based neural networks'' make it possible to study the effects of and interactions between different types of sparsities for functional and performant networks trained to do language modeling.
\cite{Neil2017} and \cite{Subramoney2023} expect that the two sparsities are orthogonal compression methods with respect to task performance, and \cite{Gao2022} demonstrate this effect, but without a detailed analysis.

In this work, we perform a detailed investigation of the synergies between activity sparsity and connectivity sparsity in recurrent networks for language modeling using a recently published event-based neural network~\cite{Subramoney2023}.
Is there a cost involved in jointly applying activity sparsity and connectivity sparsity?
To answer this question, we conduct a comparative study between densely activated Long-Short Term Memory (LSTM) networks \cite{Hochreiter1997} and sparsely activated event-based GRU (EGRU) networks \cite{Subramoney2023}. 
LSTM and EGRU both outperform conventional GRUs on these tasks.
Our results show that the two sparsities are indeed independent for the case of our EGRU language model in a large part of the design space with respect to task performance.
In the process of investigating the activity sparsity of the EGRU model, 
we furthermore uncover a mechanism that allows us to trade weight regularization for sparse activations.

%% file: content/related-work.tex
\section{Related Work}
\label{sec:related-work}
\minisec{Activity sparsity}
Spiking neural networks (SNNs) process information with discrete unary messages between neurons.
As such sparse network activity is a key feature of SNNs.
SNNs have found a wide range of applications in the machine learning domains of vision \cite{Liu2022, Fang2023}, audio \cite{hammouamri2024learning, Bittar2022}, and robotics \cite{zhao2020neuromorphic}.
Yet, biologically plausible SNNs have not yet achieved state-of-the-art performance in important machine learning tasks like language modeling.

The class of event-based neural networks relaxes the constraint of biological plausibility to improve the task performance of SNNs.
Early work in this direction showed that operating neural networks only on changes of a signal can be equivalent to operating on the original signal \cite{connor2017sigma, Neil2017}. 
Woźniak et al.~\cite{Wozniak2020} connected principles from SNNs with ANNs to reduce neuron-to-neuron communication.
However, their model only achieved perplexity \num{108} on the Penn Treebank dataset \cite{Penn1993}, where lower is better.
Rezaabad et al.~\cite{Rezaabad2020} proposed a spiking variant of Long-Short Term Memory (LSTM) networks \cite{Hochreiter1997}, where the LSTM gates were computed by SNNs with binary activations, and the state was updated through a multiplicative interaction between these binary values.
They reported a perplexity of \num{90.2} on the WikiText-2 dataset \cite{merity2017pointer}.
Along similar lines, Subramoney et al.\cite{Subramoney2023} proposed an event-based variant of the Gated Recurrent Unit (GRU) \cite{Cho2014}, which they call EGRU.
Their model achieved competitive performance with the LSTM based state-of-the-art set by \cite{Merity2018} on both Penn Treebank and WikiText-2.
We reproduced and slightly improved upon their results, leading to perplexities of \num{56.6} on Penn Treebank and \num{66.6} on WikiText-2 as reported in \secref{results}.  
Larger scale language modeling was addressed by \cite{Zhu2023}, who combined the recurrent RWKV language model \cite{peng2023rwkv} with SNNs.
Shen et al.~\cite{shen2023astrocyteenabled} trained linear transformers with spike activations of up to \num{1.5} billion parameters and got noteworthy results on large scale generative pre-training and zero-shot learning.
\minisec{Connectivity sparsity} 
The process of removing connections from neural networks is called pruning.
An extensive review of pruning techniques can be found in \cite{Hoefler2021}.
In the context of recurrent sequence models,
pruning has been applied to a range of recurrent architectures including Elman RNNs, LSTMs, and GRUs \cite{Narang2017, Han2017, Zhu2018, Bellec2018}.
On speech recognition benchmarks compression rates of up to \SI{90}{\percent} ha achieved with the LSTM model without loss in performance \cite{Han2017, Zhu2018, Bellec2018}.
Dai et al.~\cite{Dai2020} reported an increase of sparsity by expanding the linear transformations of LSTM gates to be multi layer neural networks.
The best pruned LSTM model for language modeling on the Penn Treebank dataset in the literature was reported by \cite{Zhu2018}. 
They achieved their best results, a perplexity of \num{77.5} (where lower is better), at a weight sparsity of \SI{80}{\percent}. 
We generally found that pruning has not been applied to more recent LSTM based language models such as the AWD-LSTM \cite{Merity2018}, which achieves a perplexity of \num{57.3}.
\minisec{Joint activity sparsity and connectivity sparsity} 
Weight pruning is a popular research topic in the context of SNNs \cite{Chen2018,Rathi2019, Nguyen2021, Chen2021, Kim2022, yin2023, shi2024towards}.
However, most work focused on feed-forward models for image classification.
Recurrent models are particularly relevant for non-static tasks such as audio processing or language modeling, 
but the literature on sparse connectivity remains scarce.
One example is Chakraborty et al.\cite{chakraborty2024sparse}, who proposed a pruning method for recurrent SNNs.
They evaluated on static images as well as dynamic prediction tasks such as stock market prediction.
In the context of deep learning,
Hunter et al.~\cite{Hunter2022} introduced a structured sparse algorithm for top-k winner-takes-it-all activation sparsity for feed-forward networks. 
Closest to our work is Gao et al.~\cite{Gao2022}, who achieved a significant reduction in operations by jointly applying pruning and delta-coding without compromising on their speech recognition task performance.
Furthermore, they showed that the reduction in operations translated into real efficiency gains for an FPGA-based sparse DeltaLSTM accelerator.
While they showed compelling results for both reduced operations and task performance, their focus was not on dissecting the effects of the two sparsities.
\minisec{Language modeling with RNNs}
Small scale language modeling datasets such as Penn Treebank \cite{Penn1993} or WikiText-2 \cite{merity2017pointer} drove progress of LSTM based language models before \cite{Vaswani2017} enabled large-scale language modeling.
The best raw LSTM language models on the Penn Treebank dataset are reported in \cite{Merity2018} and \cite{melis2018on}.
Recently, a class of RNNs based on linear recurrences has demonstrated strong results on large-scale language modeling tasks \cite{Gu2022, Fu2023, peng2023rwkv, gu2023mamba}.
In this work, we work in the small data regime, and comparison with the newer architectures will be done in future work.

%% file: content/methods.tex
\section{Methods}
In this section, we present the EGRU \cite{Subramoney2023}, 
and discuss pruning strategies below.We study the tradeoff between sparsities and task performance using the EGRU~\cite{Subramoney2023} architecture.
Our choice of the EGRU architecture is motivated by the fact that it is the only event-based architecture that, to our knowledge, achieves performance competitive with LSTMs on small-scale language modeling tasks.
We describe the EGRU architecture and the pruning strategies we use below.

\subsection{Event-based Gated Recurrent Unit}
Long Short-Term Memory (LSTM) networks \cite{Hochreiter1997} and Gated Recurrent Units (GRU) \cite{Cho2014} allow long range learning by gating input and hidden state variables for better gradient propagation.
The Event-based GRU (EGRU)~\cite{Subramoney2023} combines a biologically inspired spiking mechanism with the GRU gating mechanisms.
Therefore, it distinguishes between a local cell state $\bfc = (c_1, \dots, c_n)$ similar to the membrane potential in bio-plausible models, and a communicated cell state $\bfy = (y_1, \dots, y_n)$, where $n$ is the hidden dimension.
In each time-step, the communicated state $\bfy$ contains non-zero elements only where the local state $\bfc$ is larger than a learned threshold $\bfth = (\vartheta_1, \dots, \vartheta_n)$.
For $i=1,\dots,n$, $\bfy$ is given by 
\begin{align}
    \label{eq:y}
    y_i\dti \;=\; c_i\dti \, \fun{H}{ c_i\dti - \vartheta_i } \,, \quad \fun{H}{x} = \begin{cases} 1, \quad x \geq 0 \\ 0, \quad x<0 \end{cases} \,.
\end{align}
The sparse state $\bfy$ is then passed to an update gate $\bfu$ and a reset gate $\bfr$, similar to the GRU model
\begin{align}
    \bfu\dti &= \fun{\sigma}{ \bfW_{ux} \bfx\dti + \bfW_{uy} \bfy\dtim1 + \bfb_u }\,, \label{eq:u}\\
    \bfr\dti &= \fun{\sigma}{\bfW_{rx} \bfx\dti + \bfW_{ry} \bfy\dtim1 + \bfb_r } \,, \label{eq:r}
\end{align}
where $\sigma$ is the sigmoid function.
The gates compute a proposed state $\bfz$ and the updated local state $\bfc$ as
\begin{align}
     \bfz\dti &= \fun{g}{ \bfW_{zx} \bfx\dti + \bfW_{zy} \left(\bfr\dti \hp\; \bfy\dtim1\right) + \bfb_z }\,, \label{eq:z}\\
     \bfc\dti &= \bfu\dti \hp\; \bfz\dti  + (1-\bfu\dti) \hp\; \bfc\dtim1 - \bfs\dti\, , \label{eq:c}
\end{align}
where $\bfs\dti = \bfth \fun{H}{\bfc\dti - \bfth}$ is a reset term motivated by the membrane potential reset upon spikes commonly used in SNNs (see Eshragian et al.~\cite{Eshraghian2023} for a review).
The equations \eqref{eq:y} and \eqref{eq:c} are the spike and reset mechanism known from SNNs,
with the difference that equation~\eqref{eq:y} refers to a non-binary real-valued (graded) spike, i.e. $\bfy$ is not restricted to be a binary vector.
The equations \eqref{eq:u}-\eqref{eq:z} are the GRU gates.
Surrogate gradients are attached to the non-differentiable Heaviside function 
\(
    \frac{\mathrm{d}H}{\mathrm{d}c} = \lambda ~ \max
    \left(1 - \lvert c \rvert / \epsilon\right)
\)
 to allow differentiation of the event-based system, as is common practice.

\subsection{Sparsely Connected Networks}
\label{sec:pruning}
Event-based systems such as EGRU improve efficiency by reducing the activity on each neuron-to-neuron channel.
An orthogonal method to reduce the communication of a system is to remove neuron-to-neuron channels entirely, i.e. prune weights of the neural network (see sec. \ref{sec:related-work}).
The most popular heuristic for weight removal is weight magnitude pruning \cite{Han2015}.
In weight magnitude pruning, weights with the smallest magnitudes are systematically identified from a chosen set of target tensors and  a specified percentage is removed by setting it to zero.
Following the recommendations in \cite{Hoefler2021}, we investigated several pruning routines. 
We used a two-step approach, where we first trained the RNN model to convergence followed by sparsifying it through iterative pruning, which produced the best results for our goals of inference performance and sparsity.

The specific pruning methodology we implemented was a global unstructured weight magnitude pruning technique. 
At each step, we carried out weight magnitude pruning on all the weight tensors that constituted the RNN model. 
The weights to be pruned were selected globally from all the tensors except for the embedding vectors. 
By selecting the weights globally, we enabled the layers that play a larger role in the forward pass to retain a commensurate proportion of its weights.
Our rationale for pruning the RNN weights and not the embeddings was based on the perception that RNN weights are more representative across different tasks rather than just language modeling.
After each pruning iteration, we allowed the model to fine-tune for a few epochs before advancing to the subsequent pruning step. 
This iterative procedure was repeated until a pre-defined target sparsity level was achieved.

\subsection{Efficiency of Sparse Activations and Sparse Connectivity}
The efficiency gains of sparse activations and sparse weights complement each other in a multiplicative way.
This yields significantly more efficient systems compared to ones that have each of these sparsities separately.
Consider the linear transformation $\bfW \bfa$.
Let $\lambda_\text{a} = \mathbb{E}\left[ \bfa \neq 0 \right]$ denote the fraction of active neurons in each time step.
We then call $\sigma_\text{a} = 1 - \lambda_\text{a}$ the activation sparsity.
Likewise, we call $\sigma_\text{w}$ the weight sparsity of $\bfW$, and denote the fraction of non-zero connections $\lambda_\text{w}$.
The transformation $\bfW\bfa$ requires memory access and computation for the non-zero weights $\bfW_{ij}$ for each non-zero $\bfa_j$.
Hence, we need to load $\lambda_\text{a}$ columns of $\bfW$ that each have a fraction of $\lambda_\text{w}$ non-zero values. 
Effectively the fraction of remaining operations compared to a dense vector matrix multiplication is $\lambda_\text{a} \cdot \lambda_\text{w}$.

%% file: content/results.tex
\section{Results}
\label{sec:results}
We conducted our experiments on the Penn Treebank \cite{Penn1993} and the WikiText-2 \cite{merity2017pointer} small-scale language modeling tasks.
Both tasks are word-level language modeling tasks, 
where Penn Treebank consists of about \num{1} million words, with a dictionary of \num{10000} unique words, 
and WikiText-2 features selected Wikipedia articles of about \num{2.5} million words, with a dictionary of \num{33278} unique words.
The task performance metric for both datasets was measured in perplexity, i.e. the exponential of cross-entropy.
As such, lower perplexities indicate better performance.
Since computational efficiency depends on hardware properties, there is no universal metric to quantify efficiency.
Bio-plausible models are commonly compared in terms of synaptic operations \cite{yik2024neurobench}.
The equivalent metric for EGRU, which incurs multiplications, are multiply accumulate (MAC) operations.
Here, we chose MACs as a fine-grained measure of theoretically required operations on digital hardware.

All models trained for this work followed the architecture of \cite{Subramoney2023}, which was based on \cite{Merity2018}.
An embedding look-up table for the word embeddings was followed by three layers of stacked RNNs without skip connections and a linear decoder, whose weights were tied to the embedding layer.
DropConnect was applied to the recurrent weights \cite{Wan2013}.
In contrast to \cite{Subramoney2023, Merity2018}, we significantly simplified the optimization procedure by using AdamW instead of their proposed averaged SGD schedule.
AdamW speeds up the convergence of both LSTM and EGRU by a factor of 3-4 in comparison to Adam or ASGD \cite{Merity2018}.
While it would be natural to choose GRU as a baseline for comparison with EGRU, GRU models did not match the LSTM performance in our experiments.
This is consistent with the literature that does not report GRU results close to the LSTM baseline.

A comparison of the state-of-the-art recurrent neural networks with our models on the Penn Treebank and WikiText-2 dataset is presented in \tabref{ptb-sota} and \tabref{wt2-sota}, respectively.
These results only include models that did not use additional text data for pre-training.
As the only model employing both activity sparsity and weight sparsity, our pruned EGRU achieves competitive results compared to dense LSTM baselines \cite{melis2018on, Merity2018}.
Our model clearly outperforms the only other model with activity sparsity \cite{Rezaabad2020}, who report results only on WikiText-2,
as well as the models of \cite{Zhu2018} with sparse weights.
Since our work focuses on the interaction of activity sparsity and weight sparsity, we did not use orthogonal strategies such as mixture-of-softmaxes \cite{yang2018breaking}, neural cache \cite{Grave2017}, or mogrifier gates \cite{Melis2020Mogrifier} that can further improve the performance of recurrent language models.
\begin{table}[]
    \centering
    \caption{
        State-of-the-art comparison of recurrent language models solely trained on the Penn Treebank dataset \cite{Penn1993} without extra training data.
        Task performance was measured in perplexity (PPL), where lower is better.
        We report the bare recurrent network results for fair comparison without dynamic evaluation or Monte-Carlo dropout sampling.
    }
    \begin{tabular}{l c S[table-format=2.0] S[table-format=2.1]}
    \toprule
         \multirow{2}{*}{Model} & Activity & {Weight} & {\multirow{2}{*}{Test PPL $\downarrow$}}  \\
         & sparsity & {sparsity} & \\
    \midrule
         LSTM \cite{Zhu2018} & \ding{55} & \SI{80}{\percent} & 83.9 \\
         LSTM \cite{Zhu2018} & \ding{55} & \SI{95}{\percent} & 96.3 \\
         LSTM \cite{melis2018on} & \ding{55} & {\ding{55}} &  58.3 \\
         AWD-LSTM \cite{Merity2018} & \ding{55} & {\ding{55}} & 57.3 \\
         EGRU \cite{Subramoney2023} & \ding{51} & {\ding{55}} & 57.2 \\
         Mogrifyier LSTM \cite{Melis2020Mogrifier} & \ding{55} & {\ding{55}} & 51.0 \\
    \midrule
        LSTM (ours) & \ding{55} & \SI{80}{\percent} & 57.6 \\
        EGRU (ours) & \ding{51} & \SI{80}{\percent} & 58.0 \\
        LSTM (ours) & \ding{55} & \SI{95}{\percent} & 66.5 \\
        EGRU (ours) & \ding{51} & \SI{95}{\percent} & 65.2 \\
    \bottomrule
    \end{tabular}
    \label{tab:ptb-sota}
\end{table}
\begin{table}[]
    \centering
    \caption{
        State-of-the-art comparison of recurrent language models solely trained on the WikiText-2 dataset \cite{merity2017pointer} without extra training data.
        Task performance was measured in perplexity (PPL), where lower is better.
        We report the bare recurrent network results for fair comparison without dynamic evaluation or Monte-Carlo dropout sampling.
        Mixture-of-Softmax models are denoted by MoS. }
    \begin{tabular}{l c S[table-format=2.0] S[table-format=2.1]}
    \toprule
         \multirow{2}{*}{Model} & Activity & {Weight} & {\multirow{2}{*}{Test PPL $\downarrow$}}  \\
         & sparsity & {sparsity} & \\  
     \midrule
         LSTM SNN \cite{Rezaabad2020} & \ding{51} & {\ding{55}} & 91.2 \\
         EGRU \cite{Subramoney2023} & \ding{51} & {\ding{55}} & 70.6 \\
         AWD-LSTM \cite{Merity2018} & \ding{55} & {\ding{55}} &  65.8 \\
         AWD-LSTM (MoS) \cite{yang2018breaking} & \ding{55} & {\ding{55}} &  61.5 \\
         Mogrifyier LSTM (MoS) \cite{Melis2020Mogrifier} & \ding{55} & {\ding{55}} & 56.6 \\
    \midrule
        LSTM (ours) & \ding{55} & \SI{80}{\percent} & 68.0 \\
        EGRU (ours) & \ding{51} & \SI{80}{\percent} & 69.4 \\
        LSTM (ours) & \ding{55} & \SI{95}{\percent} & 81.9 \\
        EGRU (ours) & \ding{51} & \SI{95}{\percent} & 79.3 \\
    \bottomrule
    \end{tabular}
    \label{tab:wt2-sota}
\end{table}
\subsection{Joint activity sparsity and connectivity sparsity}
\begin{figure}
    \centering
    \includegraphics[width=\linewidth]{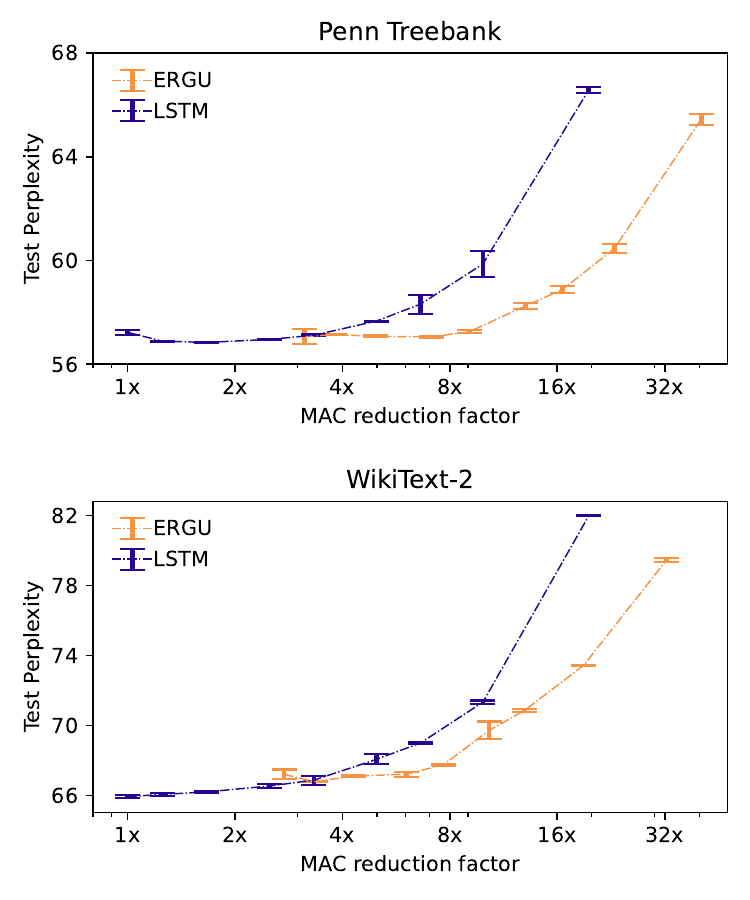}
    \caption{
        Weight sparsity (pruned LSTM) vs joint activity sparsity and weight sparsity (pruned EGRU). 
        Each point corresponds to either LSTM or EGRU with a increasingly sparse connections from left to right.
        Both models show a similar performance degradation as connections are removed.
        Mean test perplexity and corresponding standard deviation over 15 random seeds.
        The detailed numbers for the best models are presented in \tabref{ptb-results} and \tabref{wikitext-results}.
    }
    \label{fig:perplexity}
\end{figure}
To understand how event-based communication interacts with sparse connectivity,
we pruned LSTM and EGRU baselines to connection sparsities between \SI{20}{\percent} and \SI{95}{\percent}. 
In particular, we wanted to test the hypothesis raised in \cite{Neil2017, Subramoney2023} that the effects of sparse connections and sparse activations are independent means of reducing operations.
If this holds in practice, 
the overall reduction in operations for a given task performance should be the product of the reduction by connection sparsity and the reduction by activity sparsity.
Considering task performance is essential, since, if the task performance would suffer stronger from jointly applying sparse activations and sparse connections (EGRU) than just removing connections (LSTM), the two methods would not be independent means of reducing computation for a given target task performance.
The amount of activity sparsity in EGRU is learned by gradient descent, in contrast to Delta Networks \cite{Neil2017}, 
so it might happen that EGRU compensates for fewer connections with higher activity.

Our results are presented in \figref{perplexity}.
For both datasets, we plotted the reduction in MAC operations versus the test perplexity averaged over 15 random seeds.
The LSTM with dense activations reduces MAC operations only through the removal of connections,
while EGRU reduces MAC operations through the joint effect of sparse connections and sparse activations.
For both models, the degradation of task performance follows the same qualitative characteristics as connections were removed. 
The characteristic of EGRU locates towards higher MAC reduction on the x-axis due to its inherently sparse activations.
The effects of sparse activations and sparse weights are clearly additive on the logarithmic axis,
and hence multiplicative as conjectured by \cite{Neil2017, Subramoney2023}.

\begin{figure}[htb]
    \centering
    \includegraphics[width=\linewidth]{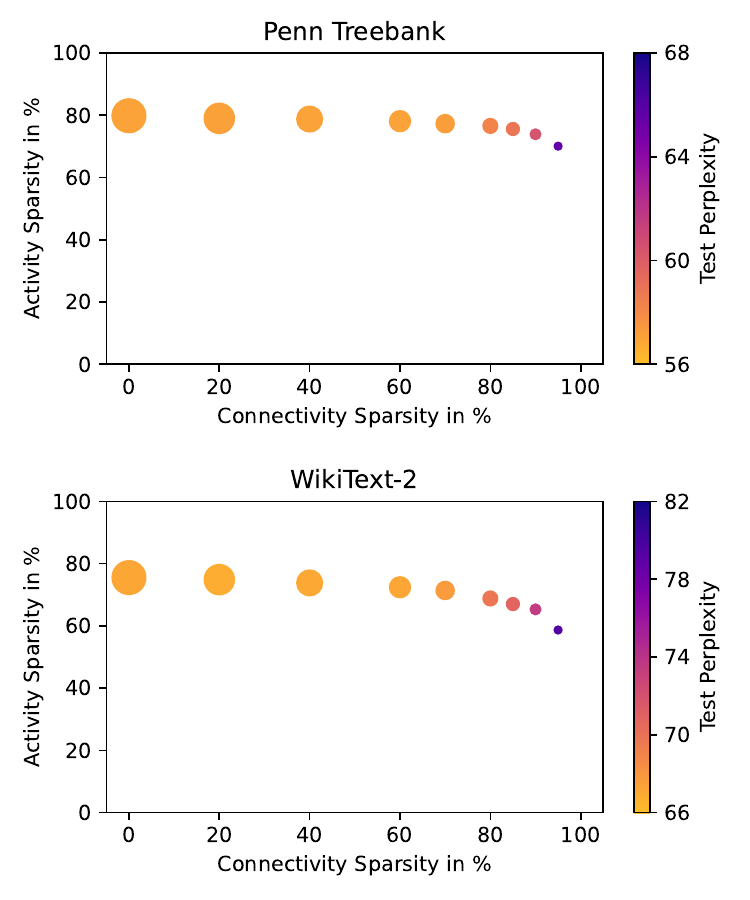}
    \caption{
        EGRU adjusts its network activity through gradient decent, which leads to different degrees of activity sparsity for corresponding degrees of connectivity sparsity.
        For a broad range of connectivity sparsity, the network activity remained almost independent.
        The training process compensates for connection sparsity with more activity in the network only when the connection sparsity was high. 
    }
    \label{fig:sparsity}
\end{figure}

In \figref{sparsity}, we show that for a wide range of connectivity sparsity the high degree of activity sparsity is maintained.
This shows that EGRU compensates for connection sparsity with higher network activity only when the connection sparsity is too high.
For both the results visualized in \figref{perplexity} and \figref{sparsity}, we observed a slightly stronger impact of pruning connections on the WikiText-2 dataset compared to the Penn Treebank dataset.
Detailed results for both datasets are shown in \tabref{ptb-results} and \tabref{wikitext-results}.
\begin{table}[htb]
\centering
\caption{
    Comparison of densely activated LSTM with sparsely activated EGRU for a range of connectivity sparsity on the Penn Treebank dataset \cite{Penn1993}.
    The test perplexity (PPL) of the model with the best validation perplexity is reported along the number of multiply-accumulate (MAC) operations for the recurrrent model only, i.e. without token read-out.
}
\begin{tabular}{c S[table-format=2.1] S[table-format=2.1] S[table-format=2.1] S[table-format=2.1]}
\toprule
\multirow{2}{*}{Weight sparsity} & \multicolumn{2}{c}{LSTM} & \multicolumn{2}{c}{EGRU} \\
\cmidrule(lr){2-3} \cmidrule(lr){4-5}
& {MAC} & {Test PPL $\downarrow$} & {MAC} & {Test PPL $\downarrow$} \\
\midrule
\SI{0}{\percent}  &    20.2M &  57.1 & 6.4M & 56.6 \\
\SI{20}{\percent}  &    16.2M &  56.9 & 5.3M & 57.1 \\
\SI{40}{\percent}  &    12.1M &  56.8 & 4.1M & 56.9 \\
\SI{60}{\percent}  &     8.1M &  56.9 & 2.8M & 57.0 \\
\SI{70}{\percent}  &     6.1M &  57.1 & 2.2M & 57.1 \\
\SI{80}{\percent}  &     4.1M &  57.6 & 1.6M & 58.0 \\
\SI{85}{\percent}  &     3.1M &  57.7 & 1.2M & 58.7 \\
\SI{90}{\percent}  &     2.0M &  58.3 & 0.9M & 60.2 \\
\SI{95}{\percent}  &     1.0M &  66.5 & 0.5M & 65.2 \\
\bottomrule
\end{tabular}
\label{tab:ptb-results}
\end{table}
\begin{table}
\caption{
    Comparison of densely activated LSTM with sparsely activated EGRU for a range of connectivity sparsity on the WikiText-2 dataset \cite{merity2017pointer}.
    The baselines are \cite{Merity2018} for the LSTM and \cite{Subramoney2023} for the EGRU.
    The test perplexity (PPL) of the model with the best validation perplexity is reported  along the number of multiply-accumulate (MAC) operations for the recurrrent model only, i.e. without token read-out..
}
\begin{tabular}{c S[table-format=2.1] S[table-format=2.1] S[table-format=2.1] S[table-format=2.1]}
\toprule
\multirow{2}{*}{Weight sparsity} & \multicolumn{2}{c}{LSTM} & \multicolumn{2}{c}{EGRU} \\
\cmidrule(lr){2-3} \cmidrule(lr){4-5}
& {MAC} & {Test PPL $\downarrow$} & {MAC} & {Test PPL $\downarrow$} \\
\midrule
\SI{0}{\percent} &  20.2M &  65.7 &  7.4M &  66.6 \\
\SI{20}{\percent}  &  16.2M &  65.9 &  6.0M &  66.7 \\
\SI{40}{\percent} &  12.1M &  66.1 &  4.7M &  67.0 \\
\SI{60}{\percent} &   8.1M &  66.4 &  3.3M &  67.1 \\
\SI{70}{\percent} &   6.1M &  66.3 &  2.6M &  67.6 \\
\SI{80}{\percent} &   4.1M &  68.0 &  2.0M &  69.4 \\
\SI{85}{\percent} &   3.1M &  68.9 &  1.6M &  70.7 \\
\SI{90}{\percent} &   2.0M &  71.2 &  1.1M &  73.3 \\
\SI{95}{\percent} &   1.0M &  81.9 &  0.6M &  79.3 \\
\bottomrule
\end{tabular}
\label{tab:wikitext-results}
\end{table}
\subsection{Activity sparsity and weight regularization}
\label{sec:activity-sparsity}

\begin{figure}[htb]
    \centering
    \includegraphics[width=0.99\linewidth]{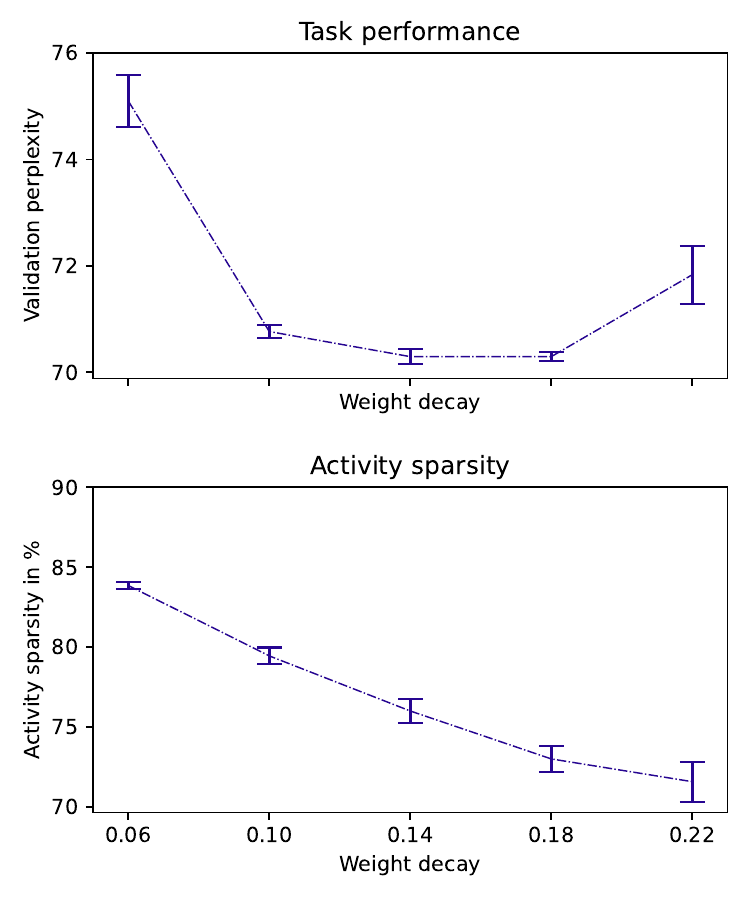}
    \caption{
        Effect of weight decay regularization on the performance and activity sparsity of the EGRU.
        We considered multiple degrees of weight decay for the weights and bias, separately.
        Means and errors are plotted for fixed decay rate on the weights and varying decay rate on biases.
        All models were trained on the larger WikiText-2 dataset \cite{merity2017pointer}.
    }
    \label{fig:trade-off}
\end{figure}
\begin{figure}[htb]
    \centering
    \includegraphics[width=0.99\linewidth]{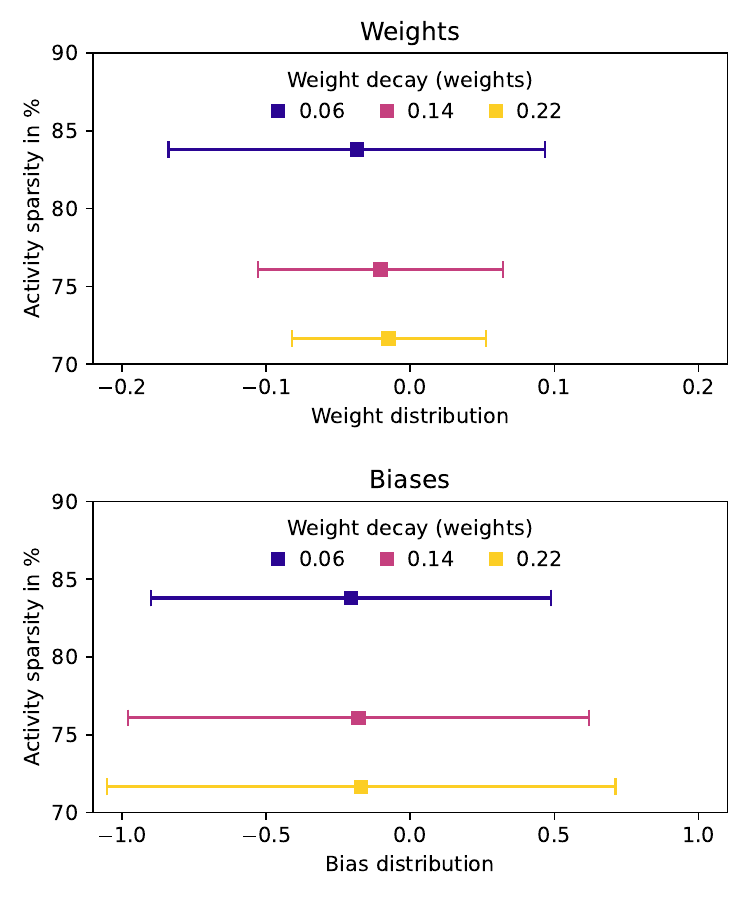}
    \caption{
        Effect of varying degrees of weight decay applied to the weights on the distribution of weights and biases for fixed weight decay applied to the biasses of \num{0.01}.
        All models were trained on the larger WikiText-2 dataset \cite{merity2017pointer}.
    }
    \label{fig:distributions}
\end{figure}
In our hyperparameter search, we observed that weight decay regularization strongly influenced both the task performance as well as the activity sparsity.
A trade-off between the regularizing effects of sparsity and weight decay emerged.
This provides a mechanism for tuning the activity of EGRU networks, which has not been observed in this context before.
In particular, this observation allowed us to trade efficiency in terms of network activity for task performance.
We systematically studied the influence of weight decay on the EGRU model by training a set of models with different degrees of weight decay separately applied to the weights and biases.

\Figref{trade-off} shows that the task performance experiences an optimum around a weight decay of \num{0.14}.
At the same time, weight decay regularization influences the network activity as shown in the bottom panel of \figref{trade-off}.
Comparing the effects of regularizing biases versus regularizing weights,
we found that regularizing weights had a stronger influence on both task performance and network activity.
Investigating the distributions of weights and biases in more detail, it became evident that both weights and biases had a tendency to be negative (\figref{distributions}).
This drove the cell states below their threshold and promoted sparse activations. 
With stronger weight decay regularization the distributions concentrated closer to zero as expected. 

%% file: content/discussion.tex
\section{Discussion}

In this study, we investigated the joint effects of sparse activity and sparse connectivity in the EGRU, a versatile event-based neural network architecture.
We provided evidence that sparse activity and connectivity are qualitatively independent directions to reduce the number of computational operations (\figref{perplexity}).
Jointly applying both strategies only affects task performance for high degrees of sparsity beyond \SI{80}{\percent} (\figref{sparsity}).
Furthermore, we dived into the learned parameters of the model and found that the training process drove the mean values of weights and biases below \num{0} (\figref{distributions}).
This fostered sparse network activity but interfered with standard weight regularization techniques such as weight decay.
We showed that network activity can be tuned with weight decay regularization to meet the requirements of a hardware system on network activity (\figref{trade-off}).
The task performance presented in \tabref{ptb-sota} and \tabref{wt2-sota} significantly improves upon previously published results of both sparsely activated RNNs \cite{Wozniak2020, Rezaabad2020} and sparsely connected RNNs \cite{Zhu2018} on the Penn Treebank and WikiText-2 benchmarks.

Implementing performant and functional machine learning models on energy-efficient neuromorphic hardware requires several design decisions that align with the engineering goals of neuromorphic systems.
To achieve meaningful task performance for language modeling, it is necessary to use network architectures designed for task performance independent of detailed biological plausibility while retaining the features of sparse communication and event-based computation.
These features are inspired by biology without being constrained by biology, and are essential for achieving energy efficiency on neuromorphic hardware.
Furthermore, due to limitations in memory and communication bandwidth  in neuromorphic hardware, it is critical to combine and use both activity and parameter sparsity.
Understanding the interaction between the two (as in this work) allows us to estimate the characteristics of such a model in terms of energy efficiency, task performance, and latency.
Overall, this provides a step towards improving existing implementations~\cite {Nazeer2024} and moving towards more complex models on neuromorphic hardware.